\documentclass[sigconf]{acmart}

\AtBeginDocument{%
  }

\copyrightyear{2026}
\acmYear{2026}
\setcopyright{cc}
\setcctype{by}
\acmConference[KDD 2026] {Proceedings of the 32nd ACM SIGKDD Conference on Knowledge Discovery and Data Mining V.2}{August 9--13, 2026}{Jeju Island, Republic of Korea.}
\acmBooktitle{Proceedings of the 32nd ACM SIGKDD Conference on Knowledge Discovery and Data Mining V.2 (KDD 2026), August 9--13, 2026, Jeju Island, Republic of Korea}
\acmISBN{979-8-4007-2259-2/2026/08}
\acmDOI{10.1145/3770855.3818446}
\usepackage{pifont}
\usepackage{enumitem}
\usepackage{multirow}
\usepackage{stfloats}
\usepackage{balance}

\newcommand{\stitle}[1]{\vspace*{0.5em}\noindent{\bf #1\/}}
\newcommand{\model}{MiCU}
\newcommand{\dataset}{DevCmd}

\author{Haowei Han}
\authornote{Work done during internship at Xiaomi Corporation.}
\authornote{Both authors contributed equally to this research.}
\orcid{0009-0002-5201-3893}
\affiliation{%
  \department{School of Computer Science}
  \institution{Wuhan University}
  \city{Wuhan}
  \country{China}
}
\email{haowei.han@whu.edu.cn}

\author{Kexin Hu}
\authornotemark[2] % 引用第一个 authornote
\orcid{0009-0008-2525-2147}
\affiliation{%
  \institution{Xiaomi Corporation}
  \city{Wuhan}
  \country{China}
}
\email{hukexin@xiaomi.com}

\author{Weiwei Cai}
\orcid{0009-0001-7346-2885}
\affiliation{%
  \institution{Xiaomi Corporation}
  \city{Wuhan}
  \country{China}
}
\email{caiweiwei@xiaomi.com}

\author{Debiao Zhang}
\orcid{0009-0009-1403-407X}
\affiliation{%
  \institution{Xiaomi Corporation}
  \city{Wuhan}
  \country{China}
}
\email{zhangdebiao@xiaomi.com}

\author{Bin Qin}
\orcid{0009-0004-1774-6041}
\affiliation{%
  \institution{Xiaomi Corporation}
  \city{Wuhan}
  \country{China}
}
\email{qinbin@xiaomi.com}

\author{Yuxiang Wang}
\orcid{0000-0002-5483-8322}
\affiliation{%
  \department{School of Computer Science}
  \institution{Wuhan University}
  \city{Wuhan}
  \country{China}
}
\email{nai.yxwang@whu.edu.cn}

\author{Jiawei Jiang}
\authornote{Corresponding authors.}
\orcid{0000-0003-0051-0046}
\affiliation{%
  \department{School of Computer Science}
  \institution{Wuhan University}
  \city{Wuhan}
  \country{China}
}
\email{jiawei.jiang@whu.edu.cn}

\author{Xiao Yan}
\orcid{0000-0002-2122-915X}
\affiliation{%
  \department{Institute for Math \& AI}
  \institution{Wuhan University}
  \city{Wuhan}
  \country{China}
}
\email{yanxiaosunny@whu.edu.cn}

\author{Bo Du}
\authornotemark[3] % 引用第二个 authornote (即 Corresponding author)
\orcid{0000-0002-0059-8458}
\affiliation{%
  \department{School of Computer Science}
  \institution{Wuhan University}
  \city{Wuhan}
  \country{China}
}
\email{dubo@whu.edu.cn}
\renewcommand{\shortauthors}{Haowei Han et al.}

\begin{document}

\title{\model: End-to-End Smart Home Command Understanding with Large Language Model}

\begin{abstract}
    Command understanding systems in smart home ecosystems can automate device control and substantially improve user experience. However, while they perform well on precise utterances (e.g., "turn on the bedroom light"), they struggle with ambiguous or misaligned commands (e.g., "make the bedroom cozy"). Large language models (LLMs) generalize well across various domains and can outperform traditional rule-based systems on such tasks, but their effectiveness is often constrained by scarce domain-specific data, insufficient task-specific adaptation, and high computational costs. 
    In this paper, we propose an automated training data synthesis workflow using user logs and LLMs; then we build \model, a domain-specific LLM that excels at command understanding. Specifically, we employ curriculum learning to inject domain knowledge into the base LLM, then we enhance its reasoning ability via cold-start training combined with reinforcement learning (RL) guided by domain-specific thinking rules. Additionally, we introduce a token compression technique that condenses device description into a single special token, substantially reducing inference overhead and enabling \model-fast, an efficient variant optimized for long inputs. 
    Extensive experiments show that \model\ significantly outperforms baselines, with an average accuracy gain of 20.01\% across all device categories. We have deployed \model\ in the Xiaomi Home app, receiving approximately 1.7 million page views per day. Production evaluations show that \model\ reduces user correction rate by 1.57\% and increases human audited accuracy by 32.05\%. Our data and code are available at \url{https://github.com/xiaomi-research/iot_spec_llm}
\end{abstract}




\maketitle

\section{Introduction}

Smart home ecosystems have fundamentally transformed daily life, providing significant convenience and becoming a crucial component in modern living. The Xiaomi Home platform, a leading smart home ecosystem in China, has developed over 113 sub-category terminal devices, such as lights, curtains, and air conditioners, covering numerous aspects of daily life. Xiaomi Home connects these devices using IoT technology and manages them through its mobile application (i.e., Xiaomi Home app). By September 2025, the Xiaomi Home ecosystem reached a major milestone as connected IoT devices surpassed the 1 billion mark. This growth is supported by a robust community of 114.6 million monthly active users, including over 21.6 million users who own five or more smart devices.\footnote{Xiaomi Corporation, \href{https://ir.mi.com/static-files/1e19fb17-dc12-4803-8426-376008c467a9}{Third Quarter 2025 Results Announcement}}

\begin{figure}
    \centering
    \includegraphics[width=\linewidth]{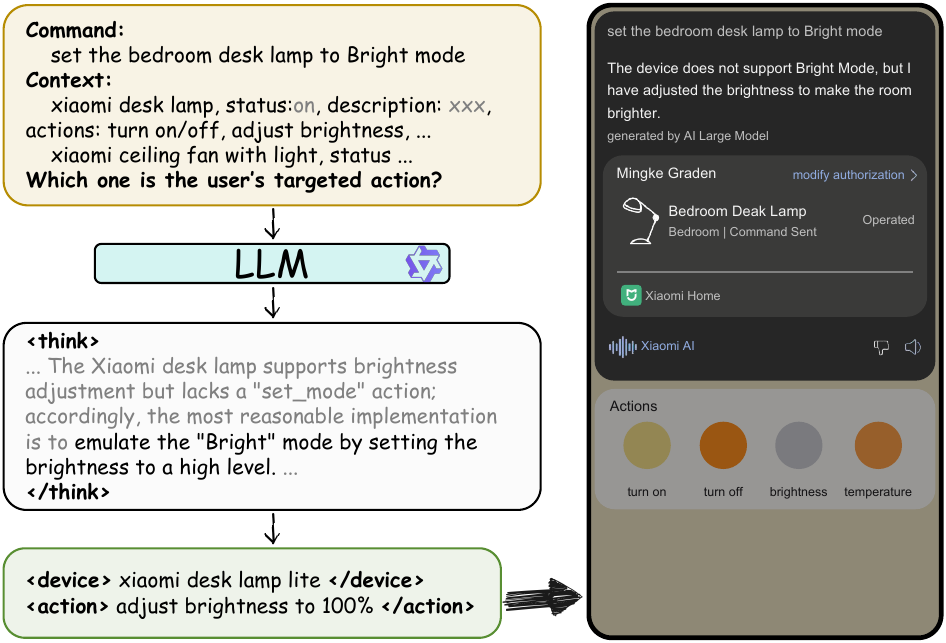}
    \caption{Illustration of the LLM-based end-to-end command understanding process in Xiaomi's smart home ecosystem.}
    \label{fig:alpha}
\end{figure}

Device control via natural language enables users to manage the Xiaomi Home ecosystem through intuitive voice commands. Given the vast diversity of devices and their complex actions, natural language interaction offers superior flexibility over rigid manual controls (e.g., remote controllers or app menus), significantly reducing user effort in multi-device management. However, achieving this capability is challenging because many user commands are ambiguous or misaligned. For example, a user may say "set the bedroom desk lamp to Bright mode" while actually intending to make the desk lamp brighter, not to select a "Bright" mode (which is not supported by the desk lamp). In such scenarios, the system must analyze the command, consider the context, and infer the user's intended action. Traditional approaches, such as rule-based selection~\cite{chill}, typically employ multi-stage heuristic pipelines that map commands to potential actions using manually designed rules (e.g., device name matching and action name matching), which increases system complexity and limits generalization and accuracy.

Large language models (LLMs), which acquire extensive world knowledge and strong reasoning ability from vast training corpora, have generalized well across various domains, including semantic parsing~\cite{roy2023benchclamp, drozdov2022compositional}, open-domain question answering~\cite{kamalloo2023evaluating, yang2023empower}, and instruction following~\cite{zhou2023instruction, lou2024large}. Prompting LLMs with user commands and context offers an intuitive, end-to-end approach to command understanding. This strategy leverages the reasoning capabilities of LLMs to effectively address the complexities of smart home interactions. As shown in Figure~\ref{fig:alpha}, LLMs utilize their strong reasoning abilities to identify the user's intent to adjust brightness higher, correctly avoiding the selection of a non-existent "Bright" mode, and showing significant advantage over traditional approaches.

\begin{table}[t]
\centering

\caption{Performance comparison among rule-based selection (RS), DeepSeek-R1 and our \model\ across all device categories in understanding smart home command.}
\begin{tabular}{lccc}
\toprule
\textbf{Metric} & \textbf{RS}   & \textbf{DeepSeek-R1}  & \textbf{\model} \\ \midrule
{Accuracy}      & 66.32         & 73.10                 & 94.61           \\
{Parameter}     & N.A.          & 671B                  & 4B              \\
\bottomrule
\end{tabular}
\label{tab:comparison}
\end{table}
\stitle{Challenges.} 
     As shown in Table~\ref{tab:comparison}, while LLM-based approaches demonstrate superior understanding compared to traditional methods, directly using general LLMs in smart home task exhibits limited performance. We identify three key challenges in developing an effective LLM-based end-to-end command understanding system:
    
    \begin{enumerate}[label=\textbf{(\arabic*)}, leftmargin=*, itemsep=0.5em]
        \item \textbf{Scarce Domain-Specific Data.} LLMs derive their capabilities from massive data. Although raw user logs in smart home scenarios are voluminous, they are difficult to use for training due to their unstructured nature and the absence of ground truth targeted actions to serve as training supervision.
        \item \textbf{Limited Task-Specific Adaptation.} While LLMs possess rich world knowledge, they lack specialized domain expertise, hence cannot well utilize context information, infer user intent and predict targeted actions. However, effective strategies for adapting LLMs specifically to the smart home command understanding task have yet to be fully developed.
        \item \textbf{High Computational Overhead.} Prompting LLMs incurs substantial computational overhead and time latency due to their massive parameter count. This challenge is further amplified in complex command understanding task, which necessitates the inclusion of a long context prompt detailing device status and other descriptions such as available actions.
    \end{enumerate}

To address \textbf{Challenge (1)}, we introduce \textit{\dataset}, a difficulty-graded dataset synthesized based on user logs. Specifically, we propose an automated synthesis workflow to generate both easy and hard command data. For easy data, the workflow populates predefined templates using device specifications which define their functional capabilities and constraints, to generate samples and their corresponding target actions as labels, following a structured format. For hard data, the workflow samples and formats hard real user logs and then leverages LLMs with retrieval augmented generation (RAG)~\cite{lewis2020retrieval} and SelfRefine~\cite{madaan2023self} to infer user intent and annotate the targeted action as the label. The resulting dataset contains 50K samples across 28 categories, effectively mitigating the scarcity of training data in smart home command understanding.

We then build \textit{\model}, an LLM for command understanding, to address the remaining challenges. For \textbf{Challenge (2)}, instead of relying on manual prompt engineering which incurs significant context overhead, we propose an effective training framework to adapt base LLM for command understanding. We first inject specialized knowledge into the base LLM via curriculum learning~\cite{xu2020curriculum}, where the training starts with the easy samples and progressively advances to the hard ones. Furthermore, we guide DeepSeek-R1~\cite{guo2025deepseek} to generate chain of thought (CoT) processes to solve the command understanding task and use these CoT processes to conduct supervised fine-tuning (SFT) of \model. Finally, we apply reinforcement learning (RL) to further refine and strengthen the model’s reasoning abilities.
For \textbf{Challenge (3)}, we identify that large redundancy originates in lengthy prompts used to describe candidate devices, such as available actions. Therefore, we use a token compression technique to condense these long description into a special token, significantly reducing the computational costs and enabling \model-fast, which retains competitive command understanding performance.

We evaluate our \model\ on datasets with multiple device categories, and compare it with baselines including traditional rule-based selection and popular generally pretrained LLMs with few-shot contextual learning. Results show that \model\ outperforms all baselines across various device categories: it achieves accuracy gains of 28.29\% and 20.01\% over the rule‑based method and the best performing LLM, respectively. Ablation experiments confirm that each stage of our training framework is effective. The efficiency study shows that token compression reduces prompt length by 31.9\% with a slight accuracy loss of only 0.6\%. Online experiments show that \model\ achieves a 1.57\% decrease in user correction rate and a 32.05\% improvement in human audited accuracy.

In this paper, our contributions are summarized as follows:

\begin{itemize}[leftmargin=*]
    \item  We introduce \dataset, a dataset of 50K training samples synthesized from real user logs, to address the data scarcity problem for smart home command understanding.
    \item  We propose \model, a specialized LLM for smart home command understanding. It integrates curriculum learning, CoT SFT and RL to inject domain knowledge and enhance reasoning ability.
    \item We design a token compression technique to significantly reduce the computational cost and latency by condensing lengthy device descriptions into single special tokens.
    \item Extensive experiments conducted in both offline and online settings across various device categories demonstrate the effectiveness of the proposed dataset synthesis workflow and \model.
\end{itemize}

\begin{figure}
    \centering
    \includegraphics[width=\linewidth]{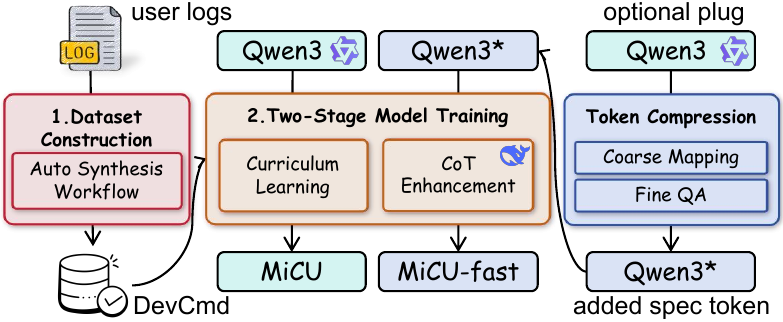}
    \caption{Overview of the key techniques of \model.}
    \label{fig:overview}
\end{figure}

\section{\model\ for Command Understanding}

Figure~\ref{fig:overview} depicts an overview of building our smart home command understanding LLM, \model. We first synthesize domain‑specific training dataset, \dataset, via an automated workflow, then adapt a base LLM (i.e., Qwen3-4B-instruct~\cite{yang2025qwen3}) using a two-stage training strategy. Furthermore, we propose an optional token compression to lower computational cost, yielding a fast variant, \model-fast. The following sections describe each component in detail.

\subsection{Dataset Construction}
\label{sec:dataset}

    To mitigate the scarcity of smart home command training data, we collected a large volume of user logs from the Xiaomi Home platform. This data collection process is conducted with the explicit informed consent of users and strictly adheres to privacy and security protocols to ensure all sensitive information is anonymized. However, this raw data is inherently unformatted, characterized by inconsistent metadata and heterogeneous structures across different device versions. More importantly, the logs are often noisy, as the inherent ambiguity of real-world scenarios makes it difficult to identify genuine user intent, thereby complicating the extraction of reliable ground truth labels for model supervision.
    
    \begin{figure}
        \centering
        \includegraphics[width=\linewidth]{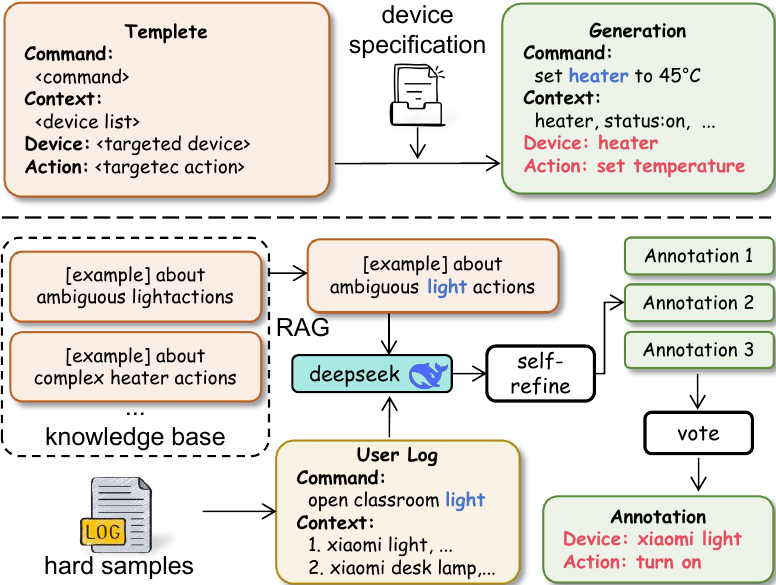}
        \caption{Dataset synthesis workflow. The top shows direct template-based generation for easy data, while the bottom shows advanced annotation for hard data using DeepSeek.}
        \label{fig:dataset}
    \end{figure}
    
    \stitle{Synthesis Workflow.}
        To construct clear, formatted training data with reliable labels, we propose an automated data synthesis workflow. This process leverages LLMs to automatically generate training samples based on real user logs. As shown in Figure~\ref{fig:dataset}, to achieve higher synthetic data quality and better adaptation to downstream training, we employ different generation strategies to create both "easy" and "hard" data samples. This dual-strategy approach yields \dataset\ $\mathcal{D}$, a difficulty-graded dataset specifically tailored for training LLMs in smart home command understanding task. As shown by the generation output in Figure~\ref{fig:dataset}, each sample in $\mathcal{D}$ comprises a user command, context, and ground truth target device and action. 
    
        For generating the easy samples $\mathcal{D}_{\text{domain}}^{\text{easy}}$, we utilize a template-based synthesis approach designed to produce samples following the required logic and the target structured output format. We systematically populate carefully predefined templates with every category of device specifications, thereby generating a corresponding simple sample for each valid action while ensuring functional completeness. This strategy enables us to efficiently gather a large volume of samples automatically, ensuring they inherently possess ground truth labels. This comprehensive coverage facilitates the robust initial training of the downstream models.
    
        For generating the hard samples $\mathcal{D}_{\text{domain}}^{\text{hard}}$, we directly adopt complex user logs, format them, and automatically annotate them with high quality labels. \ding{192} To achieve this, we manually annotate a few samples with a CoT style thinking process to construct a knowledge base of nearly 1K entries.
        Then, we leverage a RAG~\cite{lewis2020retrieval} approach to automatically annotate the rest of the samples. Then, we use each sample as a query to retrieve relevant manually annotated samples as few-shot examples, prompting DeepSeek-R1~\cite{guo2025deepseek} to infer the targeted action as label, in a CoT manner.
        \ding{193} To further ensure the quality of the synthetic labels, we employ SelfRefine~\cite{madaan2023self}, prompting the model to reflect on its CoT annotation process and attempt to regenerate it. \ding{194} We collect multiple results generated through reflection and leverage the model to vote on them, using the consensus as the ground truth label. Finally, we manually review and correct the labels annotated by the LLM, ensuring their reliability.
        This robust, multi-stage approach effectively leverages the LLM's reasoning capability to minimize the manual annotation cost while ensuring the generation of high quality labels.
    
    \stitle{Dataset Details.}
        For training, our dataset \dataset\ comprises 30K easy samples and 20K hard samples, totaling 50K samples, and covers 28 categories of smart devices, primarily lighting fixtures, but also curtains and air conditioners (AC). For evaluation, the dataset includes 9K manually annotated samples from user logs to ensure label quality and enable more precise model evaluation. Details about device categories are provided in Appendix~\ref{appendix:dataset}.

\begin{figure}
    \centering
    \includegraphics[width=\linewidth]{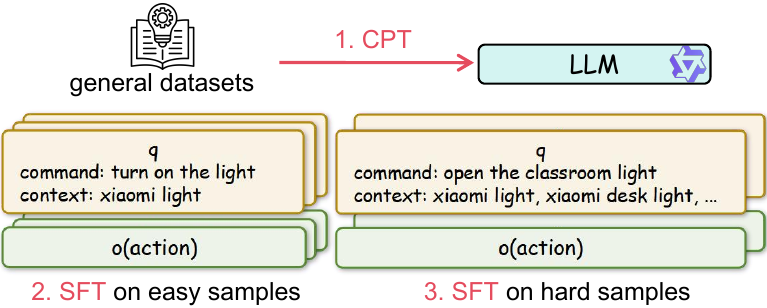}
    \caption{Curriculum learning stage. It begins with continual pre-training (CPT) of the base LLM on general datasets (e.g., WuDao), followed by progressive supervised fine-tuning (SFT) on synthesized easy-to-hard samples.}
    \label{fig:curriculum learning}
\end{figure}

\subsection{Curriculum Learning}
    While general LLMs excel across various tasks, they often struggle with specialized domains due to a lack of targeted knowledge. In smart home command understanding, task difficulty varies significantly, ranging from simple single device controls to ambiguous multi-intent commands. Directly training the model on datasets with such high difficulty variance can lead to instability and degraded performance~\cite{xu2020curriculum}. To address this, we adopt a curriculum learning approach, which introduces knowledge progressively as shown in Figure~\ref{fig:curriculum learning}. We begin with continued pre-training (CPT) on general data, followed by SFT on easy samples to establish a stable foundation, and finally SFT on challenging hard samples to ensure robust handling of complex queries.

    \stitle{General CPT.}
        To enhance the general language understanding and comprehension capabilities of the base model, we perform CPT on general datasets prior to domain-specific learning. The training objective is to minimize the negative log-likelihood (NLL), and the CPT loss is formally defined as follows:
        \begin{equation}
            \mathcal{L}_{\text{CPT}}(\mathcal{D}; \theta) = -\frac{1}{|\mathcal{D}|} \sum_{x \in \mathcal{D}} \sum_{t=1}^{|x|} \log P(x^t | x^{<t}; \theta),
        \end{equation}   
        where $\mathcal{D}$ represents the training dataset, $\theta$ denotes the model parameters, $x\in\mathcal{D}$ is a single sample, $x^t$ is the $t$-th token in the sequence, and $x^{<t}$ indicates all preceding tokens in the sequence.
        Specifically, we first train the model on two open-source Chinese corpora, Wudao~\cite{zhao2022wudaocorpora} and Firefly~\cite{Firefly}, to learn general linguistic and contextual representations, and then train it on LongAlign~\cite{bai-etal-2024-longalign} to enhance its long‑context understanding, as smart home contexts are typically long. The process can be expressed as:
        \begin{equation}
            \theta_{\text{CPT}} \xleftarrow{} \arg\min_{\theta} \mathcal{L}_{\text{CPT}}(\mathcal{D}_{\text{general}}; \theta_{\text{base}}).
        \end{equation}

    \stitle{Domain-Specific SFT.}
        We perform SFT on a continued pre-trained model to improve its domain‑specific capabilities. During SFT, the model is given user commands and contextual information as inputs and is trained to generate the corresponding target action as output. Formally, the loss function is defined as:
        \begin{equation}
            \mathcal{L}_{\text{SFT}}(\mathcal{D};\theta) = -\frac{1}{|\mathcal{D}|} \sum_{(q,o) \in \mathcal{D}} \sum_{t=1}^{|o|}  \log P(o^t | q, o^{<t}; \theta),
            \label{equ:sft}
        \end{equation}
        where $q$ denotes the input prompt, which includes both the command and context information, while $o$ represents the model's output, corresponding to the target action $a$ here.
        However, high difficulty variance and severe imbalance between easy and hard samples hinder stable training~\cite{xu2023wizardlm}. Specifically, hard samples are crucial yet complex and often scarce. Training on hard samples before easy ones destabilizes the model due to the lack of a solid foundation. Although mixing the two types is feasible, it degrades performance because the model often overfits to abundant simple patterns and fails to learn from the scarce hard ones.
        
        To address these challenges, we apply progressive SFT. We first fine-tune the model on easy samples from \dataset\ to establish foundational knowledge covering all valid device actions. This step equips the model with a comprehensive understanding of basic smart home commands and device behavior. Subsequently, we fine-tune the model on the hard samples from \dataset, which contain complex commands such as 
        multiple or ambiguous intentions, and challenging contexts like misleading device aliases. This second stage enables the model to handle complex scenarios effectively and significantly enhances its command comprehension capabilities. The fine-tuning process can be expressed as follows:
        \begin{equation}
            \theta_{\text{SFT}}' \xleftarrow{} \arg\min_{\theta} \mathcal{L}_{\text{SFT}}(\mathcal{D}_{\text{domain}}^{\text{easy}}; \theta_{\text{CPT}}),
        \end{equation}
        \begin{equation}
            \theta_{\text{SFT}} \xleftarrow{} \arg\min_{\theta} \mathcal{L}_{\text{SFT}}(\mathcal{D}_{\text{domain}}^{\text{hard}}; \theta_{\text{SFT}}').
        \end{equation}

\begin{figure}
    \centering
    \includegraphics[width=\linewidth]{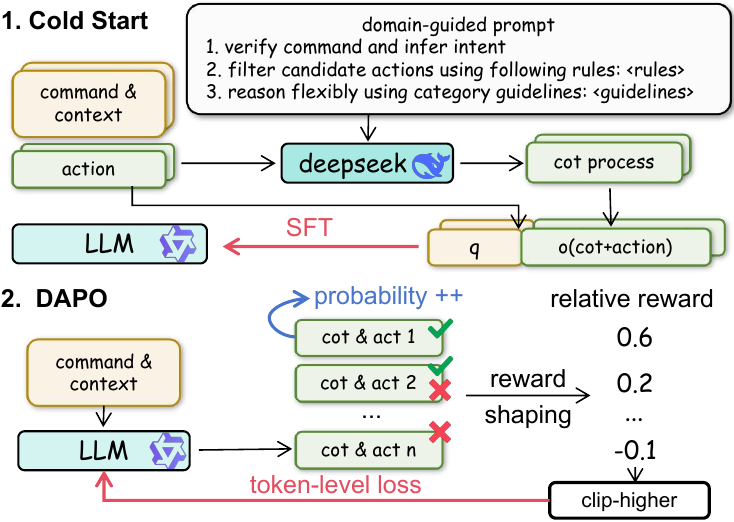}
    \caption{CoT enhancement stage. We first cold start model via SFT on domain-guided DeepSeek-generated CoT processes, and then refine it using DAPO reinforcement learning.}
    \label{fig:cot enhancement}
\end{figure}

\subsection{CoT Enhancement}
    Large language models (LLMs) have demonstrated the ability to generate improved outputs by leveraging their reasoning capabilities. Recent research has primarily focused on employing this ability through the chain-of-thought (CoT) approach, which encourages models to reason step by step. However, allowing LLMs to autonomously determine their reasoning process neglects the use of domain-specific knowledge, leading to reduced accuracy in specialized tasks.
    To address this limitation, we propose a domain-guided prompt that explicitly guides our model to align its reasoning process with domain-specific rules. Specifically, as shown in Figure~\ref{fig:cot enhancement}, we first employ the domain-guided prompt to guide a teacher, DeepSeek-R1~\cite{shao2024deepseekmath}, to generate reasoning processes for each sample, denoted as $think$. We provide targeted actions to the teacher LLM to achieve higher precision in reasoning. Then we conduct SFT on our model using the generated processes $think$ as a cold start and leverage reinforcement learning (RL) to further enhance its reasoning ability, ultimately improving the model's performance.

    \stitle{Domain-Guided Prompt.}
        As shown in Figure~\ref{fig:cot enhancement}, the reasoning process guided by the domain-guided prompt consists of three steps. First, the model verifies the validity of the user command and analyzes its intent, such as identifying which device category the user wants to operate. Second, the model applies a rule-based filtering process to narrow down the scope of potential actions. This process involves matching device names, identifying room names, and employing other filtering mechanisms aligned with predefined rules. By integrating the LLM's advanced reasoning capabilities with carefully designed business rules, this step achieves higher accuracy than traditional approach. Finally, if the previous steps fail to determine the target action, the model is instructed to flexibly reason about the command using category-specific guidelines. By employing this domain-guided prompt, large-scale teacher LLMs can produce more consistent and accurate CoT reasoning processes, significantly enhancing the training of our model.
        
    \stitle{CoT Cold Start.}
        We cold start our model with CoT reasoning ability through SFT. Specifically, the command and its corresponding context (without the domain-guided prompt) are used as the input, and the model is trained via SFT to emulate the pre-generated reasoning processes $think$ and predict the target action $a$. Building on the previous training stage, we integrate this stage into curriculum learning, which encompasses both easy and hard samples. A total of 17K CoT reasoning processes are generated for the easy samples, compared with 8K for the hard samples.
        The SFT training loss is defined in Eq.~\eqref{equ:sft}, here the output $o$ includes not only the targeted action $a$ but also the thinking process $think$ (i.e., $o=\{think,a\}$). We denote the dataset containing CoT thinking processes as $\mathcal{D}_{\text{CoT}}$. The process can be formally expressed as:
        \begin{equation}
            \theta_{\text{CoT}} \xleftarrow{} \arg\min_{\theta} \mathcal{L}_{\text{SFT}}(\mathcal{D}_{\text{CoT}}; \theta_{\text{SFT}}).
        \end{equation}

    \stitle{RL with DAPO.}  
        After training the model with a CoT cold start, which equips it with a strong reasoning foundation following the guided process, we further enhance its reasoning ability through RL by leveraging DAPO~\cite{yu2025dapo}, a large-scale RL strategy built upon GRPO~\cite{shao2024deepseekmath}. In short, the key idea of DAPO is to have the model generate multiple outputs for the same input prompt, compute the relative advantage of each output, and reward well-formed and accurate ones to increase their probability. The objective to optimize the model as a policy is as follows:
        \begin{align}
            \mathcal{J}_{\text{DAPO}}(\mathcal{D};\theta) = & \mathbb{E}_{(q, a) \sim \mathcal{D}, \{o_i\}_{i=1}^{G} \sim \pi_{\theta_{\text{old}}}(\cdot|q)} \\
            \Bigg[
                \frac{1}{\sum_{i=1}^{G}\!|o_i|}
                \sum_{i=1}^{G} \sum_{t=1}^{|o_i|}
                \min & \!\Big( r_{i}^t(\theta) \hat{A}_{i}^t, 
                \text{clip} \big( r_{i}^t(\theta), 1-\epsilon_{\text{l}}, 1+\epsilon_{\text{h}} \big) \hat{A}_{i}^t \Big)
            \Bigg] \notag \\
            \text{s.t.} & \quad 0 < |\{o_i  |  \text{is\_equivalent}(a, o_i)\}| < G, \notag 
        \end{align}
        where $G$ denotes the number of sampled outputs, $\epsilon_l$ and $\epsilon_h$ are the lower and upper bounds for clipping, respectively, $\pi_{\theta}$ and $\pi_{\text{old}}$ denote the current policy and the old policy, respectively, $r_{i}^t(\theta)$ represents token-wise importance weight for the output $o_{i}^t$ under the current policy, and $\hat{A}_{i}^t$ denotes the normalized advantage function:
        \begin{align}
            r_{i,t}(\theta) = \frac{\pi_{\theta} \big(o_{i}^t \mid q, o_{i}^{<t} \big)}{\pi_{\text{old}} \big(o_{i}^t \mid q, o_{i}^{<t} \big)} , \quad
            \hat{A}_{i,t} = \frac{R_i - \text{mean}(\{R_i\}_{i=1}^G)}{\text{std}(\{R_i\}_{i=1}^G)}.
        \end{align}
        Compared to GRPO, DAPO employs clip-higher, dynamic sampling, token-level loss, and overlong reward shaping tricks to improve the RL process, providing more stable training and higher accuracy.

        We design the reward $R_i$ to consist of three components including response tag checking, format validation checking, and content correctness checking, to ensure that the model produces correct answers in both content and format. Response tag checking encourages the model to reason and predict correctly between the designated tags <think> and </think>, <device> and </device>, <action> and </action>. Format validation checking verifies that the model’s structured output (JSON) is syntactically valid and conforms to the expected fields and types. Content correctness checking assesses the correctness of predictions, including device, action type and corresponding value. These components are weighted as needed to balance format strictness and comprehension accuracy. Overall, we can formally express the RL process as follows: 
        \begin{equation}
            \theta_{\text{ \model }} \xleftarrow{} \arg\max_{\theta} \mathcal{J}_{\text{DAPO}}(\mathcal{D}_{\text{domain}}; \theta_{\text{CoT}}).
        \end{equation}

\begin{figure}
    \centering
    \includegraphics[width=\linewidth]{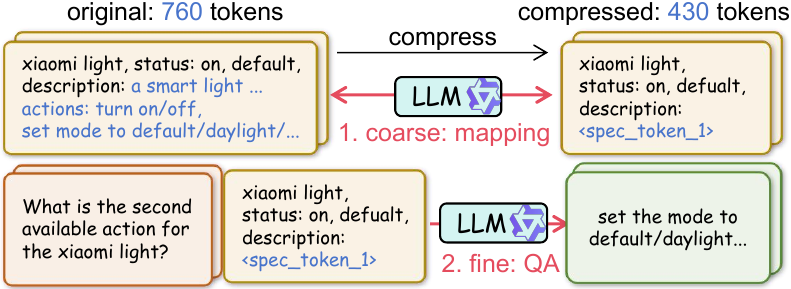}
    \caption{Token compression technique. The model undergoes a coarse-to-fine adaptation, progressing from the bi-directional mapping of description-special token pairs to task-specific QA fine-tuning.}
    \label{fig:token compression}
\end{figure}
\subsection{Token Compression}
    Although \model, trained on specialized data, effectively addresses the smart command understanding task, it incurs substantial computational overhead and latency. Our profiling reveals that the average context prompt contains 353 tokens, significantly surpassing the average output of 52 tokens. Notably, device descriptions, which detail device characteristics and permissible actions, account for 94\% of the total context prompt length. These descriptions incur substantial computational overhead, a burden that escalates linearly as the number of user devices grows. However, omitting such descriptions impairs the model’s ability to learn comprehensive device information and leads to degraded understanding. Therefore, we apply token compression technique to compress these descriptions and accelerate our \model, yielding its efficient variant, \model-fast.

    \stitle{Special Token Design.}
        The descriptions of device categories serve as detailed specifications, enabling the use of token compression techniques to represent each lengthy category specification as a unique token $t_{\text{spec}}$. To achieve this, we propose a strategy to transform verbose specifications into compact tokens. As shown in Figure~\ref{fig:token compression}, for each device category, all available actions and their corresponding value ranges are compressed into a single $t_{\text{spec}}$. For instance, a lengthy text description enumerating multiple functional capabilities (e.g., adjust brightness) and valid parameter constraints (e.g., 0-100) is replaced by a representative token, such as <spec\_token\_1>. Following this compression, a device is represented solely by its $t_{\text{spec}}$ and its simplified status, which characterizes the current state of all operational attributes. This replacement of exhaustive specifications with compact tokens significantly reduces the total token count, thereby enhancing encoding efficiency while reducing computational overhead and inference latency.
    
    \stitle{Adaptation Training.}
        To ensure the model effectively comprehends and utilizes these special tokens, we propose a coarse-to-fine training approach for adaptation as shown in Figure~\ref{fig:token compression}. First, we train the model on a bi-directional mapping task: given a special token $t_{\text{spec}}$ as input, the model is trained to generate its corresponding specification . Conversely, given a category specification, the model is trained to predict the associated $t_{\text{spec}}$. This bi-directional mapping allows the model to quickly adapt to newly added tokens, ensuring robust comprehension of tokenized specifications. Next, we train the model on a more fine-grained task: we query the model about the details of the current token $t_{\text{spec}}$, and the model is trained to generate the correct action name and its valid value range. This fine-grained task ensures the model understands each special token thoroughly, thereby achieving accurate and efficient downstream command understanding. Notably, this compression stage is incorporated before the two-stage model training so that the model is trained on sufficiently token-compressed data, enhancing its adaptability and effectiveness for downstream tasks.

\section{Experimental Evaluation}
\begin{table*}[!t]
\centering
\caption{Command understanding evaluation. Accuracy (\%) is reported. The best results are marked with bold. "Light" denotes all the devices of targeted actions in this sub-dataset are lights. Gain is the improvement over the best-performing baseline.}
\begin{tabular}{lccccc|ccccc|c}
\toprule
                                    & \multicolumn{5}{c|}{\textbf{Device Identification}} & \multicolumn{5}{c|}{\textbf{Action Prediction}}                         \\ \cmidrule(lr){2-6} \cmidrule(lr){7-11}
\multirow{-2}{*}{\textbf{Method}}   &Light &AC    &Fan   &Curtain &\textbf{Whole} &Light &AC    &Switch &Socket &\textbf{Whole} &\multirow{-2}{*}{\textbf{Overall}} \\ \midrule
Rule-based Selection                &61.37 &72.52 &74.19 &71.92   &76.98          &73.00 &30.00 &57.02  &60.87  &45.53          &66.32                              \\ \midrule
\multicolumn{12}{c}{Proprietary LLM} \\ \midrule
GPT-4o-mini                         &33.97 &52.01 &51.55 &34.13   &71.04          &45.37 &55.49 &31.40  &50.31  &49.66          &63.79                              \\
GPT-4o                              &66.99 &57.51 &62.37 &40.08   &80.43          &50.79 &67.84 &41.32  &62.11  &60.19          &73.57                              \\ 
DeepSeek-R1                         &72.58 &60.17 &65.09 &46.67   &79.51          &49.89 &68.04 &36.36  &50.93  &60.60          &73.10                              \\  
DeepSeek-V3.2                       &76.65 &77.66 &74.23 &51.19   &82.46          &48.76 &65.88 &30.74  &37.27  &59.29          &74.60                              \\ \midrule
\multicolumn{12}{c}{Open Source LLM} \\ \midrule
Llama3.1-8B                         &41.95 &38.97 &39.66 &23.17   &63.65          &36.79 &37.84 &34.13  &27.33  &36.09          &54.31                              \\  
Llama3.3-70B                        &45.99 &43.25 &45.18 &29.90   &70.03          &40.18 &49.91 &38.55  &36.08  &45.18          &61.61                              \\  
Qwen3-4B                            &31.19 &10.60 &28.88 &13.65   &39.19          &29.57 &17.84 &28.93  &36.65  &24.21          &34.11                              \\  
Qwen3-30B                           &51.52 &42.18 &49.25 &29.99   &72.85          &42.25 &51.80 &38.55  &39.50  &46.79          &64.01                              \\ 
\model-4B                           &\textbf{89.81} &\textbf{94.51} &\textbf{92.27} &93.25            &\textbf{96.80} &87.03          &\textbf{92.01} &83.24           &80.49           &90.33          &\textbf{94.61}       \\
\model-4B-fast                      &88.22          &\textbf{94.51} &91.75          &\textbf{94.44}   &95.89          &\textbf{87.21} &90.78          &\textbf{88.43}  &\textbf{85.09}  &\textbf{90.35} &94.01           \\  \midrule
Gain                                &13.16          &16.85          &18.04          &43.25            &14.34          &36.42          &23.97          &47.11           &22.98           &29.75          &20.01           \\ \bottomrule
\end{tabular}
\label{tab:main}
\end{table*}

In this section, we conduct extensive experiments to address the following research questions:

\begin{itemize}[leftmargin=*]
    \item RQ1: What overall accuracy does our model \model\ achieve on the dataset \dataset\ compared to baseline methods?
    \item RQ2: How effective are the curriculum learning stage and the CoT enhancement stage in the training framework?
    \item RQ3: How efficiently does the token compression technique improve the performance of our model \model?
    \item RQ4: Do we need more or less training data/model parameters?
    \item RQ5: How does our model \model\ perform in online deployment?
\end{itemize}

\subsection{Experimental Settings}
    \stitle{Datasets.}
    For fine-grained evaluation, we partition the evaluation set into multiple subsets based on target device categories. We select and show the most frequent and important device categories for both tasks. The training and evaluation statistics are described in Section~\ref{sec:dataset}, and further details are provided in Appendix~\ref{appendix:dataset}.
    
    \stitle{Tasks.}
        We report the accuracy performance of each method on evaluation data for two tasks with different emphases:

        \begin{itemize}[leftmargin=*, itemsep=0em]
            \item \textbf{Device Identification (DevIdent).} This task evaluates, given a command and a simple list of candidate devices, whether the model can identify the user’s target device (excluding the action). 
            \item \textbf{Action Prediction (ActPred).} This task evaluates, given a command and candidate devices with their full descriptions including available actions, whether the model can predict the user’s intended action and associated value (e.g., "set mode" and "bright"). 
        \end{itemize}
        
    \stitle{Baselines.}
        We compare \model\ and \model-fast, against three baseline types: 1) \textbf{Traditional Baseline:} rule-based selection (RS) employs a rule-based workflow to filter targeted actions by matching attributes such as device name, device property, and room name. 2) \textbf{Proprietary LLM}. We evaluate powerful proprietary LLMs including GPT-4o-mini, GPT-4o~\cite{hurst2024gpt}, DeepSeek-R1~\cite{guo2025deepseek}, and DeepSeek-V3~\cite{liu2025deepseek}. 3) \textbf{Open Source LLM}. We also assess popular open source LLMs including Llama~\cite{grattafiori2024llama} and our base model Qwen3~\cite{yang2025qwen3}. Notably, all LLM-based baselines are provided with few-shot exemplars and detailed guideline to ensure a fair comparison. Other specialized methods are excluded as they are inapplicable to this complex task.

\subsection{Main Results (RQ1)}
    Table~\ref{tab:main} reports accuracy comparisons on the evaluation data among traditional baselines, LLM-based baselines, \model\ and its efficiency variant \model-fast. We have following three observations: 

    \model\ and \model-fast consistently outperform all baselines in both DevIdent and ActPred tasks. By leveraging LLMs' superior reasoning, \model\ achieves a 28.29\% overall gain over RS. Despite its 4B parameters, \model\ surpasses much larger LLM baselines, improving by 20.01\% over DeepSeek-V3 and 30.00\% over Qwen3-30B, by effectively learning task-specific knowledge and incorporating domain-guided CoT reasoning within our training framework.
    
    Among our models, while \model-fast slightly trails \model\ in DevIdent and Overall tasks, it excels in ActPred. This is because ActPred provides full device descriptions—the primary target of our compression; here, special tokens effectively replace lengthy descriptions. Conversely, for DevIdent, the simplified descriptions cause the special tokens in \model-fast to introduce task irrelevant redundancy. We further analyze this trade-off in Section~\ref{sec:efficiency study}.

    Across both tasks, models perform better on DevIdent than ActPred, as the latter requires predicting exact actions and corresponding values, thereby making it more challenging. For device category subsets, accuracy for the "Light" category is often lower because there are frequently multiple similar candidate devices that complicate identification, and lights support a wide array of valid actions, which expands the action space and increases confusion.

\subsection{Ablation Study (RQ2)}
\label{sec:ablation}
    To study the impact of each component in the training stage, we conduct comprehensive ablation studies for them independently.

    \begin{table}[t]
\centering
\setlength{\tabcolsep}{3.5pt}
\caption{Ablation study for curriculum learning. Overall accuracy (\%) is reported with best values bolded. Hard-Easy trains the model on easy samples first, then hard samples. Mixture combines easy and hard samples during training.}
\begin{tabular}{clcccc}
\toprule
\textbf{Idx}   & \textbf{Method}              & \textbf{DevIdent} & \textbf{ActPred}  & \textbf{Overall}   \\ % 表头
\midrule
1              & Easy-only                    & 48.84             & 20.79             & 39.31              \\
2              & Hard-only                    & 94.92             & \underline{88.61} & 92.78              \\ % 内容行
3              & Easy+Hard (mixture)          & \underline{96.32} & 86.56             & \underline{93.01}  \\
4              & Hard$\rightarrow$Easy        & 68.36             & 62.49             & 66.37              \\    
5              & Easy$\rightarrow$Hard (ours) & \textbf{96.80}    & \textbf{90.33}    & \textbf{94.61}     \\ % 内容行
\bottomrule
\end{tabular}
\label{tab:curriculum learning}
\end{table}

    \stitle{Curriculum Learning.}
        As shown in Table~\ref{tab:curriculum learning}, we compare easy-only, hard-only, a mixed set, and a hard-to-easy schedule against our proposed easy-to-hard curriculum. Key observations:
        hard-only (2) outperforms easy-only (1) and achieves the second-best ActPred accuracy, indicating that hard samples contain richer information, especially for complex action prediction.
        Training on both types of samples (3) yields better DevIdent and Overall results than either type alone, demonstrating their complementarity.
        The hard-to-easy schedule (4) performs worse than the mixed setting (3), showing the training order matters. Consequently, our easy-to-hard curriculum (5) achieves the best results on both tasks.

    \begin{table}[t]
\centering
\caption{Ablation study for CoT enhancement.  Overall Accuracy (\%) is reported with best values bolded.}
\begin{tabular}{lccc}
\toprule
\textbf{Method}     & \textbf{DevIdent} & \textbf{ActPred} & \textbf{Overall}  \\ % 表头
\midrule
Base (Qwen3-4B)     & 39.19             & 24.21            & 34.11             \\
\ \ +Cold Start     & 79.86             & 67.00            & 75.50             \\ % 内容行
\ \ +RL             & 87.64             & 80.04            & 85.06             \\ \cmidrule{2-4}
Base+Curriculum     & 95.73             & 89.68            & 93.68             \\
\ \ +Cold Start     & 96.57             & 88.83            & 93.95             \\ % 内容行
\ \ +RL (ours)      & \textbf{96.80}    & \textbf{90.33}   & \textbf{94.61}    \\
\bottomrule
\end{tabular}
\label{tab:cot}
\end{table}

% \begin{table}[t]
% \centering
% \caption{Ablation study for CoT enhancement.}
% \begin{tabular}{lccc}
% \toprule
% \textbf{Method}     & \textbf{DevIdent} & \textbf{ActPred}  & \textbf{AC}       & \textbf{Overall}  \\ % 表头
% \midrule
% Base                & 27.67             & 38.31             & 47.41             & 32.47             \\
% \ \ +Cold Start     & 63.80             & 51.67             & 68.13             & 81.79             \\ % 内容行
% \ \ +RL             & 86.20             & 87.73             & 89.74             & 85.54             \\ \cmidrule{2-5}
% Base+Curriculum     & 87.58             & 89.68             & 90.84             & 95.73             \\
% \ \ +Cold Start     & 88.11             & 92.86             & 94.14             & 96.57             \\ % 内容行
% \ \ +RL             & \textbf{89.81}    & \textbf{93.25}    & \textbf{94.51}    & \textbf{96.82}    \\
% \bottomrule
% \end{tabular}
% \label{tab:cot}
% \end{table}

    \stitle{CoT Enhancement.}
        As shown in Table~\ref{tab:cot}, we evaluate model performance with and without curriculum learning under two CoT enhancement settings: cold-start and cold-start followed by reinforcement learning (RL). We observe the following.
        First, curriculum learning provides a solid foundation for CoT enhancement and consistently improves accuracy. 
        Second, CoT cold-start yields substantial gains by training the model to reason in domain-guided CoT style, which enables it to understand complex cases more effectively than directly predicting answers.
        Third, reinforcement learning further refines the model and strengthens its reasoning ability, producing additional improvements.

\begin{figure}
    \centering
    \includegraphics[width=\linewidth]{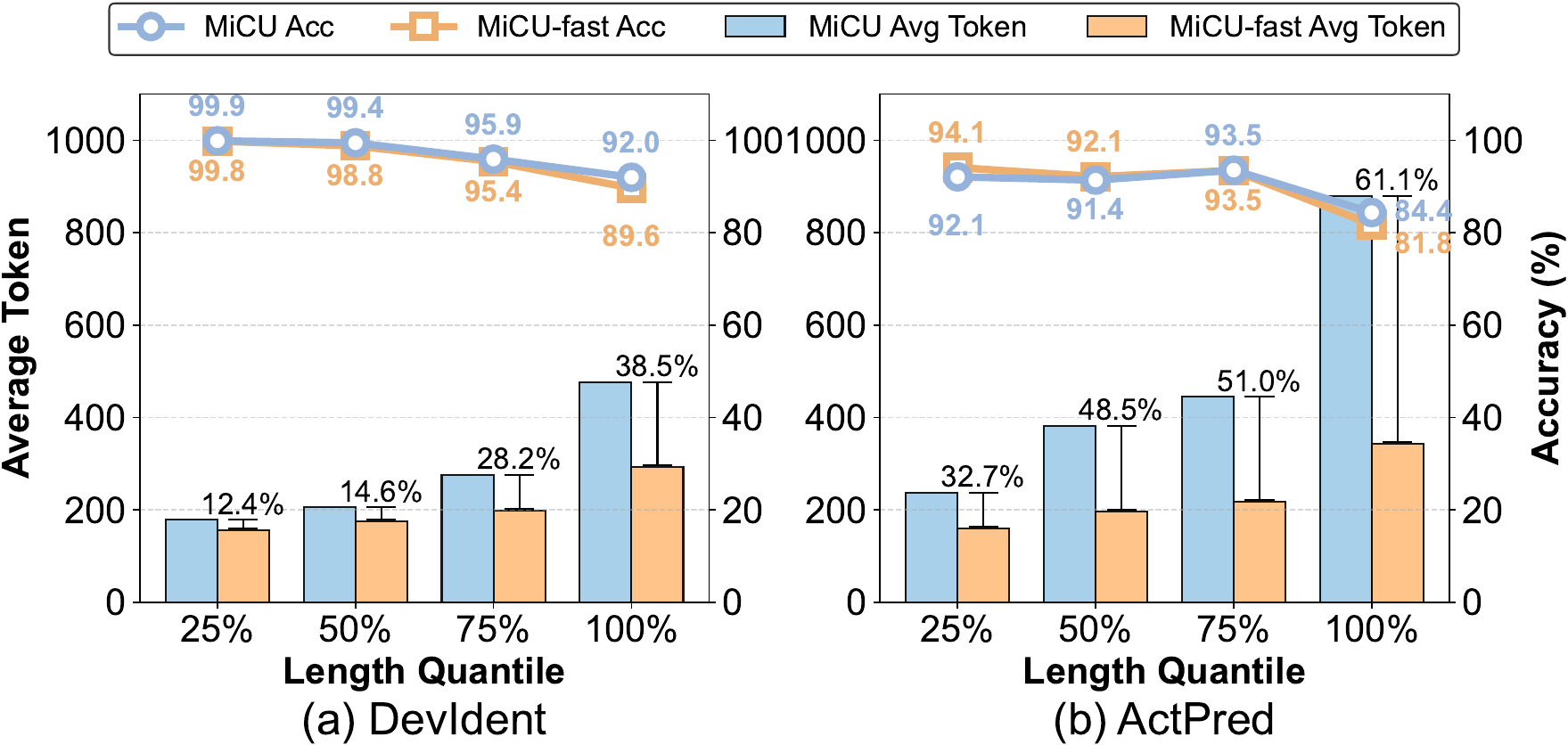}
    \caption{Efficiency study. The average token counts and corresponding accuracy for \model\ and \model-fast are shown.}
    \label{fig:efficiency}
\end{figure}

\subsection{Efficiency Study (RQ3)}
\label{sec:efficiency study}
    Figure~\ref{fig:efficiency} shows how token counts and accuracy change before and after applying our token compression technique. Key observations are as follows.
    First, accuracy inversely correlates with token count, as longer prompts introduce more candidate devices and disambiguation complexity. Consequently, the DevIdent task, which only requires device identification via simplified descriptions, yields fewer tokens and higher accuracy than the ActPred task.
    Second, \model-fast demonstrates scalable efficiency, achieving compression ratios between 12.4\% and 61.1\%. While extreme compression may cause slight accuracy loss, \model-fast notably improves accuracy on ActPred at a 48.5\% compression rate. By condensing detailed description into a single token, it filters task-irrelevant noise while preserving functional semantics. This effect is significant for ActPred's redundant descriptions but limited for DevIdent's already minimal inputs; hence, \model-fast performs relatively better on ActPred than on DevIdent.
    Overall, our technique reduces token overhead by 31.9\% and boosts QPS by 13.5\%, achieving a favorable trade-off between efficiency and performance.

\begin{figure}
    \centering
    \includegraphics[width=\linewidth]{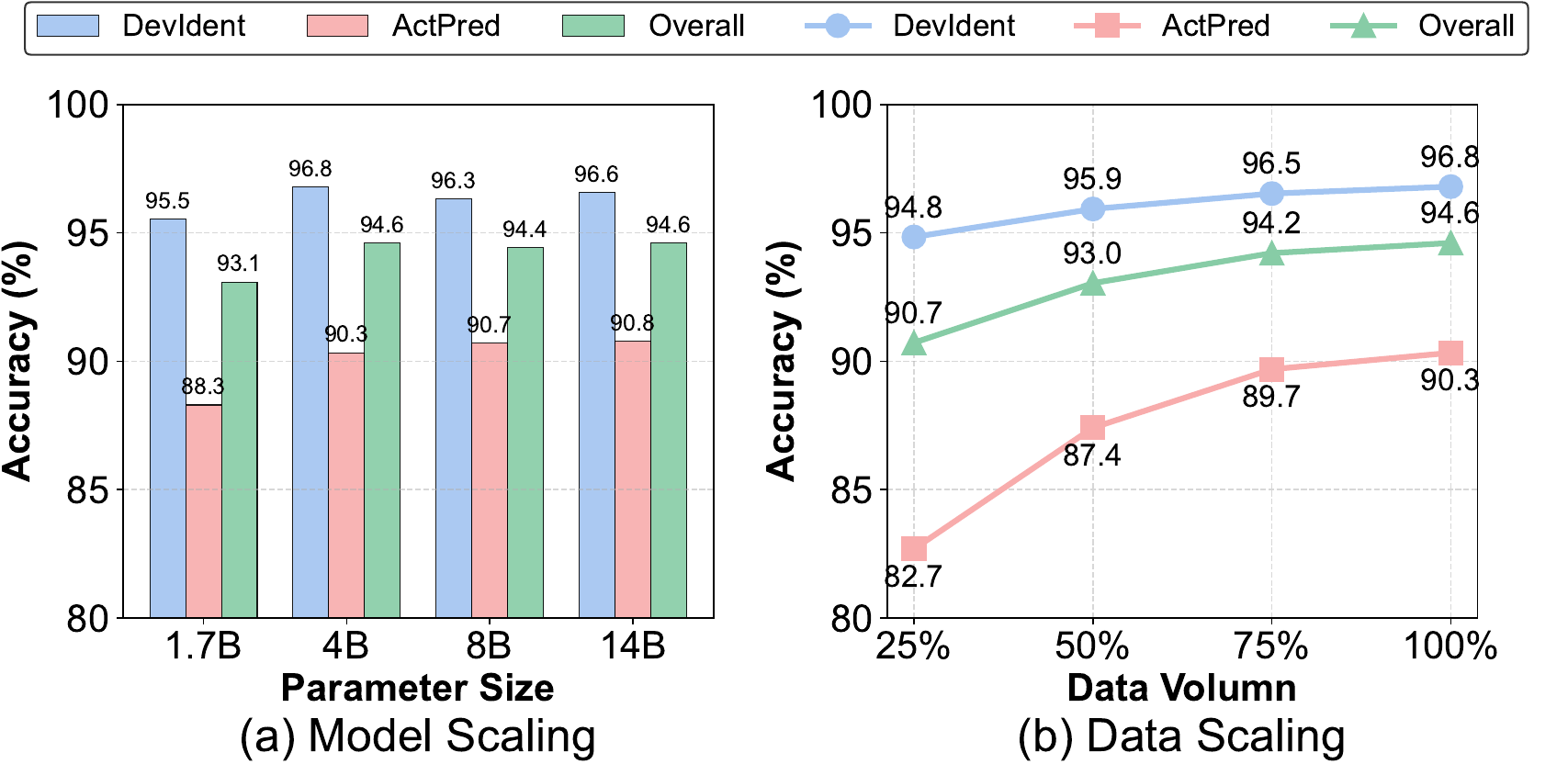}
    \caption{Scalability study. Figure (a) shows performance across base model (Qwen3) size, while figure (b) shows performance across training data size using Qwen3-4B.}
    \vspace{-8pt}
    \label{fig:scaling}
\end{figure}

\subsection{Scaling Study (RQ4)}
    We conduct scaling study by examining the effects of parameter count and training-data volume, as shown in Figure~\ref{fig:scaling}. 
    We evaluate Qwen3 models of various sizes as base models. The results indicate that very small models (e.g., 1.7B) are capacity limited and underperform, while increasing model size beyond 4B (to 8B and 14B) yields no appreciable gains in the ActPred task and actually degrades performance in the DevIdent task. Under our effective training strategy, a 4B-parameter base model is sufficient to excel in the smart home command understanding task.
    We also evaluate the effect of training data scale by proportionally increasing examples across difficulty levels and categories. For DevIdent, performance nearly plateaus once 50\%–75\% of the data is used; expanding to 100\% yields only marginal gains. For ActPred, performance remains relatively flat in the 75\%–100\% range, which we attribute to the greater complexity of this task and its need for more samples to be learned fully. Taken together, these results indicate that, for the smart home command understanding task, our current dataset is sufficient to ensure training convergence.

\subsection{Online Deployment (RQ5)}
    We deployed the proposed \model\ in Xiaomi Home’s smart home command understanding system using four instances of RTX4090 (24GiB), each provisioned with 40 GiB of memory, serving roughly 1.7 million page views (PVs) per day. The deployed service now handles the full smart home command understanding traffic. Under a sustained load of 98 QPS, the service maintains an average latency of 173 ms and a P95 latency of 263 ms, satisfying the real-time requirements for smart home interactions.
    
    \stitle{User Correction.}   
    We compared \model\ with the production baseline, a rule-based system that matches device actions using predefined heuristics (e.g., device names, properties, and room names). To quantify performance in a live environment, where ground truth labels are inherently absent, we utilized the user correction rate (UCR) as our primary online metric. UCR measures the frequency with which users manually correct the system's predicted actions, serving as a direct proxy for the failure rate of intent prediction. During a one-week A/B test, where \model\ was deployed to handle 50\% of the production traffic, results demonstrate that \model\ significantly reduces the UCR from 4.16\% to 2.59\%. This 37.37\% relative reduction over the baseline underscores the superior reliability and understanding capabilities of our model in real-world scenarios.

    \stitle{Manual Audit.} To further validate performance on real-world commands, we conducted a manual audit on sampled cases retrieved from production logs. Specifically, we randomly sampled 570 complex sessions per branch, prioritizing high frequency core categories such as lighting and air conditioning. We invited expert annotators to verify whether the predicted actions aligned with the users' underlying intents. To ensure high quality annotation, all samples underwent independent double-labeling, with any annotator inconsistencies resolved through a secondary review. Results indicate that \model\ significantly enhances the accuracy from 47.1\% to 78.3\%. This fine-grained review demonstrates that our model can effectively understand user commands and resolves intent ambiguities that traditional heuristic-based systems fail to address.

    \begin{figure}[t]
        \centering
        \includegraphics[width=\linewidth]{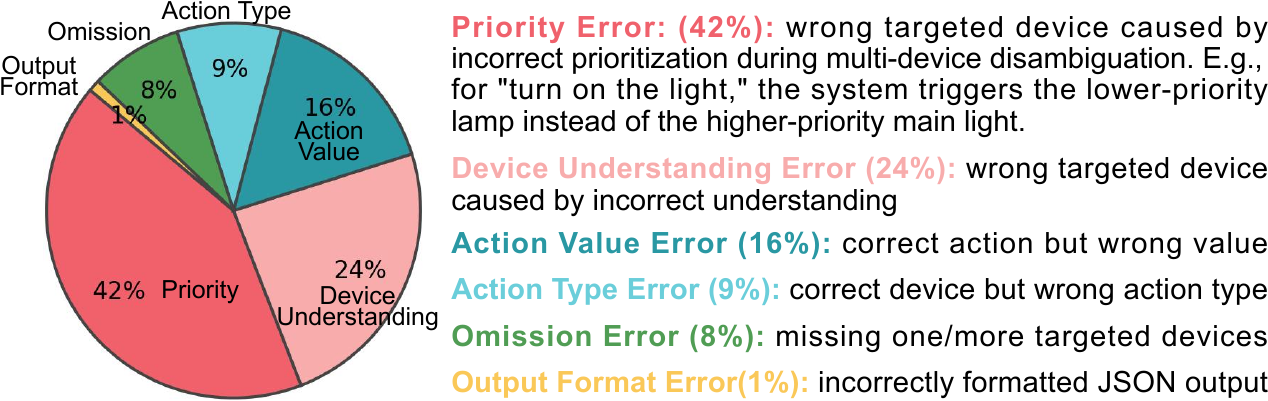}
        \caption{Failure analysis for the main category "Light".}
        \label{fig:failure analysis}
    \end{figure}

    \stitle{Failure Analysis.}
    We sample 100 failure cases for lights, which are the most common category. We categorize the failure causes into six types as shown in Figure~\ref{fig:failure analysis}. Most failures are caused by device priority errors, indicating that the complex priority relationships among devices are difficult for the model to learn. Format errors are the least frequent, suggesting our training framework effectively helps the 4B model follow instructions. The remaining failures largely stem from complex commands and contextual dependencies, which challenge the model’s command understanding capabilities.

\section{Related Work}

\stitle{LLMs for Smart Home.}
   In recent years, several works have explored applying LLMs to smart-home tasks. ChatIoT~\cite{chatiot2024} enables zero‑code trigger‑action programming (TAP) from natural language, allowing end users to create IoT automation rules via an LLM. LLM4TAP~\cite{llm4tap} infers user needs and intents from textual descriptions and combines this signal with a graph enhanced by singular value decomposition within a contrastive learning framework to improve TAP rule recommendation. SAGE~\cite{sage} grounds command understanding in concrete smart home actions by restricting generation to an LLM‑controlled action set and using a dynamically constructed prompt tree, achieving substantial gains over prior LLM baselines on a new benchmark. Sasha~\cite{sasha} addresses under‑specified, goal‑oriented commands by guiding LLMs through five structured, heuristic steps. Despite these advancements, these methods focus on TAP or goal-oriented reasoning rather than direct command understanding. Consequently, they are inapplicable to our specific task, whereas our \model\ fills this research gap.

\stitle{Command Understanding.}
    Command understanding is a long-standing challenge. Early systems~\cite{vish} relied on manual grammars and heuristic keyword extraction. Subsequent BERT-based models~\cite{dsbert, stbert, smbert} improved semantic parsing but still required heuristics for action mapping. LLMs have recently been adopted for command understanding due to their robust contextual reasoning and generalization capabilities~\cite{tellex2011understanding, silver2022pddl, xie2023translating}. Unlike traditional pipelines, LLMs excel at disambiguation, intent inference, and multi-step reasoning, enhancing end-to-end accuracy.
    To narrow the gap between general LLMs and domain‑specific command understanding, two strategies are common: prompt engineering and post-training. Prompt engineering~\cite{sahoo2024systematic, wang-etal-2022-knowledge} supplies domain knowledge and reasoning guidelines to the model at inference time via carefully designed prompts, while post‑training~\cite{posttraining, xu2023wizardlm} (e.g., domain SFT) injects domain knowledge directly into model weights. We adopt post-training to enhance performance while reducing inference overhead from lengthy prompts, echoing broader efforts in training and inference acceleration~\cite{10.1109/INFOCOM.2019.8737614, 10.1145/3770854.3783925, zheng2026efficient, zheng2026scheduling}. We further introduce automated data synthesis, two-stage training comprising curriculum learning and CoT enhancement, and a token compression technique. These innovations collectively empower \model\ with superior command understanding capabilities in the smart home domain.

\section{Conclusion}
We introduce \model, an LLM-based command understanding system for smart homes. Specifically, we use curriculum learning to inject foundational knowledge into the base LLM. We then employ CoT enhancement to improve the model's reasoning ability. Finally, we propose a token compression technique that compresses long device descriptions into single tokens, reducing the model's computational cost. Extensive offline/online experiments demonstrate that our \model\ outperforms all baselines, paving the way for community research in smart home command understanding.

\begin{acks}
We sincerely thank Fan Zhang, Siwen Wang and Kaixin Niu from Xiaomi Corporation for their constructive discussions and valuable support throughout this project.
This work was sponsored by the National Natural Science Foundation of China (No. 62225113, 62472327), the New Cornerstone Science Foundation through the XPLORER
PRIZE, the Innovative Research Group Project of Hubei Province (No. 2024AFA017), and the Key RD Program of Hubei Province (No. 2023BAB077). This work was supported by WHU-Kingsoft Joint Lab.
\end{acks}

\bibliographystyle{ACM-Reference-Format}
\balance
\bibliography{reference}

@inproceedings{roy2023benchclamp,
  author    = {Roy, Subhro and Thomson, Samuel and Chen, Tongfei and Shin, Richard and Pauls, Adam and Eisner, Jason and Van Durme, Benjamin},
  title     = {Benchclamp: A Benchmark for Evaluating Language Models on Syntactic and Semantic Parsing},
  booktitle = {Advances in Neural Information Processing Systems (NeurIPS)},
  year      = {2023}
}

@article{drozdov2022compositional,
  title={Compositional semantic parsing with large language models},
  author={Drozdov, Andrew and Sch{\"a}rli, Nathanael and Aky{\"u}rek, Ekin and Scales, Nathan and Song, Xinying and Chen, Xinyun and Bousquet, Olivier and Zhou, Denny},
  journal={arXiv preprint arXiv:2209.15003},
  year={2022}
}

@inproceedings{kamalloo2023evaluating,
  title={Evaluating open-domain question answering in the era of large language models},
  author={Kamalloo, Ehsan and Dziri, Nouha and Clarke, Charles and Rafiei, Davood},
  booktitle={Proceedings of the Annual Meeting of the Association for Computational Linguistics (ACL)},
  year={2023}
}

@inproceedings{yang2023empower,
  title={Empower large language model to perform better on industrial domain-specific question answering},
  author={Yang, Fangkai and Zhao, Pu and Wang, Zezhong and Wang, Lu and Qiao, Bo and Zhang, Jue and Garg, Mohit and Lin, Qingwei and Rajmohan, Saravan and Zhang, Dongmei},
  booktitle={Proceedings of the Conference on Empirical Methods in Natural Language Processing (EMNLP))},
  year={2023}
}

@article{zhou2023instruction,
  title={Instruction-following evaluation for large language models},
  author={Zhou, Jeffrey and Lu, Tianjian and Mishra, Swaroop and Brahma, Siddhartha and Basu, Sujoy and Luan, Yi and Zhou, Denny and Hou, Le},
  journal={arXiv preprint arXiv:2311.07911},
  year={2023}
}

@article{lou2024large,
  title={Large language model instruction following: A survey of progresses and challenges},
  author={Lou, Renze and Zhang, Kai and Yin, Wenpeng},
  journal={Computational Linguistics},
  volume={50},
  number={3},
  pages={1053--1095},
  year={2024},
}

@inproceedings{lewis2020retrieval,
  title={Retrieval-augmented generation for knowledge-intensive nlp tasks},
  author={Lewis, Patrick and Perez, Ethan and Piktus, Aleksandra and Petroni, Fabio and Karpukhin, Vladimir and Goyal, Naman and K{\"u}ttler, Heinrich and Lewis, Mike and Yih, Wen-tau and Rockt{\"a}schel, Tim and others},
  booktitle={Advances in Neural Information Processing Systems (NeurIPS)},
  year={2020}
}

@inproceedings{xu2020curriculum,
  title={Curriculum learning for natural language understanding},
  author={Xu, Benfeng and Zhang, Licheng and Mao, Zhendong and Wang, Quan and Xie, Hongtao and Zhang, Yongdong},
  booktitle={Proceedings of the Annual Meeting of the Association for Computational Linguistics (ACL)},
  year={2020}
}

@article{guo2025deepseek,
  title={Deepseek-r1: Incentivizing reasoning capability in llms via reinforcement learning},
  author={Guo, Daya and Yang, Dejian and Zhang, Haowei and Song, Junxiao and Zhang, Ruoyu and Xu, Runxin and Zhu, Qihao and Ma, Shirong and Wang, Peiyi and Bi, Xiao and others},
  journal={arXiv preprint arXiv:2501.12948},
  year={2025}
}

@article{yang2025qwen3,
  title={Qwen3 technical report},
  author={Yang, An and Li, Anfeng and Yang, Baosong and Zhang, Beichen and Hui, Binyuan and Zheng, Bo and Yu, Bowen and Gao, Chang and Huang, Chengen and Lv, Chenxu and others},
  journal={arXiv preprint arXiv:2505.09388},
  year={2025}
}

@article{hurst2024gpt,
  title={Gpt-4o system card},
  author={Hurst, Aaron and Lerer, Adam and Goucher, Adam P and Perelman, Adam and Ramesh, Aditya and Clark, Aidan and Ostrow, AJ and Welihinda, Akila and Hayes, Alan and Radford, Alec and others},
  journal={arXiv preprint arXiv:2410.21276},
  year={2024}
}

@misc{zhao2022wudaocorpora,
  author       = {Xue Zhao and Hanyu Zhao and Sha Yuan and Yequan Wang},
  title        = {WuDaoCorpora Text},
  year         = {2022},
  url          = {https://doi.org/10.57760/sciencedb.o00126.00004},
  doi          = {10.57760/sciencedb.o00126.00004},
}

@inproceedings{bai-etal-2024-longalign,
    title = "{L}ong{A}lign: A Recipe for Long Context Alignment of Large Language Models",
    author = "Bai, Yushi  and
      Lv, Xin  and
      Zhang, Jiajie  and
      He, Yuze  and
      Qi, Ji  and
      Hou, Lei  and
      Tang, Jie  and
      Dong, Yuxiao  and
      Li, Juanzi",
    booktitle = "Findings of the Association for Computational Linguistics: EMNLP",
    year = {2024},
}

@article{yu2025dapo,
  title={Dapo: An open-source llm reinforcement learning system at scale},
  author={Yu, Qiying and Zhang, Zheng and Zhu, Ruofei and Yuan, Yufeng and Zuo, Xiaochen and Yue, Yu and Dai, Weinan and Fan, Tiantian and Liu, Gaohong and Liu, Lingjun and others},
  journal={arXiv preprint arXiv:2503.14476},
  year={2025}
}

@article{shao2024deepseekmath,
  title={Deepseekmath: Pushing the limits of mathematical reasoning in open language models},
  author={Shao, Zhihong and Wang, Peiyi and Zhu, Qihao and Xu, Runxin and Song, Junxiao and Bi, Xiao and Zhang, Haowei and Zhang, Mingchuan and Li, YK and Wu, Yang and others},
  journal={arXiv preprint arXiv:2402.03300},
  year={2024}
}

@article{chatiot2024,
  author     = {Gao, Yi and Xiao, Kaijie and Li, Fu and Xu, Weifeng and Huang, Jiaming and Dong, Wei},
  title      = {ChatIoT: Zero-code Generation of Trigger-action Based IoT Programs},
  journal    = {Proceedings of the ACM on Interactive, Mobile, Wearable and Ubiquitous Technologies (IMWUT)},
  volume     = {8},
  number     = {3},
  articleno  = {103},
  year       = {2024},
}

@ARTICLE{llm4tap,
  author={Wu, Gang and Hu, Liang and Hu, Yuxiao and Xiong, Xingbo and Wang, Feng},
  journal={IEEE Internet of Things Journal}, 
  title={LLM4TAP: LLM-Enhanced TAP Rule Recommendation}, 
  year={2025},
  volume={12},
  number={10},
  pages={13157-13169},
}

@article{sage,
  title={Sage: Smart home agent with grounded execution},
  author={Rivkin, Dmitriy and Hogan, Francois and Feriani, Amal and Konar, Abhisek and Sigal, Adam and Liu, Steve and Dudek, Greg},
  journal={arXiv preprint arXiv:2311.00772},
  year={2023}
}

@article{sasha,
    author = {King, Evan and Yu, Haoxiang and Lee, Sangsu and Julien, Christine},
    title = {Sasha: Creative Goal-Oriented Reasoning in Smart Homes with Large Language Models},
    year = {2024},
    volume = {8},
    number = {1},
    journal = {Proceedings of the ACM on Interactive, Mobile, Wearable and Ubiquitous Technologies (IMWUT)},
    articleno = {12},
}

@InProceedings{vish,
author="Noura, Mahda
and Heil, Sebastian
and Gaedke, Martin",
editor="Bielikova, Maria
and Mikkonen, Tommi
and Pautasso, Cesare",
title="VISH: Does Your Smart Home Dialogue System Also Need Training Data?",
booktitle="Web Engineering",
year="2020",
pages="171--187",
}

@inproceedings{stbert,
  title={St-bert: Cross-modal language model pre-training for end-to-end spoken language understanding},
  author={Kim, Minjeong and Kim, Gyuwan and Lee, Sang-Woo and Ha, Jung-Woo},
  booktitle={IEEE International Conference on Acoustics, Speech and Signal Processing (ICASSP)},
  pages={7478--7482},
  year={2021},
}

@article{dsbert,
  title={Knowledge distillation from bert transformer to speech transformer for intent classification},
  author={Jiang, Yidi and Sharma, Bidisha and Madhavi, Maulik and Li, Haizhou},
  journal={arXiv preprint arXiv:2108.02598},
  year={2021}
}

@inproceedings{tellex2011understanding,
  title={Understanding natural language commands for robotic navigation and mobile manipulation},
  author={Tellex, Stefanie and Kollar, Thomas and Dickerson, Steven and Walter, Matthew and Banerjee, Ashis and Teller, Seth and Roy, Nicholas},
  booktitle={Proceedings of the AAAI Conference on Artificial Intelligence (AAAI)},
  year={2011}
}

@inproceedings{smbert,
  title={Semantics-aware BERT for language understanding},
  author={Zhang, Zhuosheng and Wu, Yuwei and Zhao, Hai and Li, Zuchao and Zhang, Shuailiang and Zhou, Xi and Zhou, Xiang},
  booktitle={Proceedings of the AAAI Conference on Artificial Intelligence (AAAI)},
  year={2020}
}

@article{xie2023translating,
  title={Translating natural language to planning goals with large-language models},
  author={Xie, Yaqi and Yu, Chen and Zhu, Tongyao and Bai, Jinbin and Gong, Ze and Soh, Harold},
  journal={arXiv preprint arXiv:2302.05128},
  year={2023}
}

@inproceedings{silver2022pddl,
  title={PDDL planning with pretrained large language models},
  author={Silver, Tom and Hariprasad, Varun and Shuttleworth, Reece S and Kumar, Nishanth and Lozano-P{\'e}rez, Tom{\'a}s and Kaelbling, Leslie Pack},
  booktitle={Proceedings of the NeurIPS 2022 Workshop on Foundation Models for Decision Making},
  year={2022}
}

@article{liu2025deepseek,
  title={Deepseek-v3.2: Pushing the frontier of open large language models},
  author={Liu, Aixin and Mei, Aoxue and Lin, Bangcai and Xue, Bing and Wang, Bingxuan and Xu, Bingzheng and Wu, Bochao and Zhang, Bowei and Lin, Chaofan and Dong, Chen and others},
  journal={arXiv preprint arXiv:2512.02556},
  year={2025}
}

@article{grattafiori2024llama,
  title={The llama 3 herd of models},
  author={Grattafiori, Aaron and Dubey, Abhimanyu and Jauhri, Abhinav and Pandey, Abhinav and Kadian, Abhishek and Al-Dahle, Ahmad and Letman, Aiesha and Mathur, Akhil and Schelten, Alan and Vaughan, Alex and others},
  journal={arXiv preprint arXiv:2407.21783},
  year={2024}
}

@article{madaan2023self,
  title={Self-refine: Iterative refinement with self-feedback},
  author={Madaan, Aman and Tandon, Niket and Gupta, Prakhar and Hallinan, Skyler and Gao, Luyu and Wiegreffe, Sarah and Alon, Uri and Dziri, Nouha and Prabhumoye, Shrimai and Yang, Yiming and others},
  journal={Advances in Neural Information Processing Systems (NeurIPS)},
  year={2023}
}

@misc{Firefly,
  author = {Jianxin Yang},
  title = {Firefly},
  year = {2023},
  publisher = {GitHub},
  journal = {GitHub repository},
  howpublished = {\url{https://github.com/yangjianxin1/Firefly}},
}

@article{sahoo2024systematic,
  title={A systematic survey of prompt engineering in large language models: Techniques and applications},
  author={Sahoo, Pranab and Singh, Ayush Kumar and Saha, Sriparna and Jain, Vinija and Mondal, Samrat and Chadha, Aman},
  journal={arXiv preprint arXiv:2402.07927},
  year={2024}
}

@inproceedings{wang-etal-2022-knowledge,
    title = "Knowledge Prompting in Pre-trained Language Model for Natural Language Understanding",
    author = "Wang, Jianing  and
      Huang, Wenkang  and
      Qiu, Minghui  and
      Shi, Qiuhui  and
      Wang, Hongbin  and
      Li, Xiang  and
      Gao, Ming",
    booktitle = "Proceedings of the Conference on Empirical Methods in Natural Language Processing (EMNLP)",
    year = "2022",
}

@article{posttraining,
author = {Zhang, Shengyu and Dong, Linfeng and Li, Xiaoya and Zhang, Sen and Sun, Xiaofei and Wang, Shuhe and Li, Jiwei and Hu, Runyi and Zhang, Tianwei and Wang, Guoyin and Wu, Fei},
title = {Instruction Tuning for Large Language Models: A Survey},
year = {2026},
journal = {ACM Computing Survey},
}

@inproceedings{xu2023wizardlm,
  title={Wizardlm: Empowering large language models to follow complex instructions},
  author={Xu, Can and Sun, Qingfeng and Zheng, Kai and Geng, Xiubo and Zhao, Pu and Feng, Jiazhan and Tao, Chongyang and Jiang, Daxin},
  booktitle={Proceedings of International Conference on Learning Representations (ICLR)},
  year={2023}
}

@inproceedings{chill,
author = {Zelle, John M. and Mooney, Raymond J.},
title = {Learning semantic grammars with constructive inductive logic programming},
year = {1993},
booktitle = {Proceedings of the AAAI Conference on Artificial Intelligence (AAAI)},
}

@inproceedings{10.1109/INFOCOM.2019.8737614,
author = {Hu, Chuang and Bao, Wei and Wang, Dan and Liu, Fengming},
title = {Dynamic Adaptive DNN Surgery for Inference Acceleration on the Edge},
year = {2019},
booktitle = {Proceedings of the IEEE Conference on Computer Communications (INFOCOM)},
}

@inproceedings{10.1145/3770854.3783925,
author = {Wang, Yuxiang and Ma, Chi and Yan, Xiao and Huang, Mincong and Li, Xiaoguang and Han, Ruidong and Yin, Bin and Chen, Shangyu and Li, Xiang and Jiang, Fei and Yu, Lei and Liu, Chuan and Lin, Wei and Han, Haowei and Zhou, Xiaokai and Du, Bo and Jiang, Jiawei},
title = {MTGenRec: An Efficient Distributed Training System for Generative Recommendation Models in Meituan},
year = {2026},
booktitle = {Proceedings of the ACM SIGKDD Conference on Knowledge Discovery and Data Mining (KDD)},
}

@article{zheng2026efficient,
  title={Efficient Serving for Dynamic Agent Workflows with Prediction-based KV-Cache Management},
  author={Zheng, Haoyu and Fu, Fangcheng and Wu, Jia and Yuan, Binhang and Zhang, Yongqiang and Wang, Hao and Zhu, Yuanyuan and Yan, Xiao and Jiang, Jiawei},
  journal={arXiv preprint arXiv:2605.06472},
  year={2026}
}

@article{zheng2026scheduling,
  title={Scheduling LLM Inference with Uncertainty-Aware Output Length Predictions},
  author={Zheng, Haoyu and Zhang, Yongqiang and Fu, Fangcheng and Zhou, Xiaokai and Luo, Hao and Zhu, Hongchao and Zhu, Yuanyuan and Wang, Hao and Yan, Xiao and Jiang, Jiawei},
  journal={arXiv preprint arXiv:2604.00499},
  year={2026}
}

@article{lin2026fuxishuffle,
  title={FuxiShuffle: An Adaptive and Resilient Shuffle Service for Distributed Data Processing on Alibaba Cloud},
  author={Lin, Yuhao and Tang, Zhipeng and Tong, Jiayan and Xiao, Junqing and Lu, Bin and Li, Yuhang and Li, Chao and Zhang, Zhiguo and Wang, Junhua and Luo, Hao and others},
  journal={arXiv preprint arXiv:2602.22580},
  year={2026}
}

\appendix

\section{Dataset Detail}
\label{appendix:dataset}
    The DevCmd dataset comprises 50K samples across 28 distinct smart device categories, covering a comprehensive range of modern smart home applications. These categories include environmental control devices such as air conditioners, fans, humidifiers, dehumidifiers, fresh air systems, and electric heaters, alongside sensing hardware like temperature and humidity sensors and air quality monitors. Household cleaning and maintenance are represented by robot vacuums, floor scrubbers, vacuum cleaners, washing machines, dishwashers, clothes dryers, and drying racks. Furthermore, the dataset incorporates infrastructure and lighting components including switches, sockets, smart curtains, window openers, and lamps, as well as central control panels. Finally, it covers a diverse array of entertainment and lifestyle appliances, such as televisions, split TVs, TV boxes, set-top boxes, projectors, rice cookers, water heaters, bath heaters, smart mirrors, aroma diffusers, foot baths, and smart flower pots. This extensive variety ensures the dataset captures a wide spectrum of linguistic expressions and functional commands across different domestic scenarios.

\section{Implementation}
We use Qwen3-4B-instruct~\cite{yang2025qwen3} as the base LLM and deepseek-R1~\cite{guo2025deepseek} for chain-of-thought (CoT) generation. During the CPT stage the model is trained for 1 epoch with a learning rate of 5e-5. For curriculum SFT, the easy and hard stages are trained for 1 and 2 epochs, respectively, with a learning rate of 5e-6. In the RL phase of CoT enhancement we use a learning rate of 1e-5 and set the clip ratios $\epsilon_l$ and $\epsilon_h$ to 0.2 and 0.28 for stability. Across all stages, the model is fully trained with all parameters optimized~\cite{lin2026fuxishuffle}. Training was conducted on 8 NVIDIA H20 GPUs (96GB). For online deployment, we utilized FP8 quantization and implemented speculative decoding with a lookahead step of $k=4$. This inference strategy proves highly effective for our task, as it significantly accelerates the generation of highly structured tokens in the JSON outputs.

\section{Privacy}
    Smart home command understanding relies on user commands and contextual signals (e.g., device status), which may contain sensitive information. Therefore, we follow a privacy-by-design principle throughout data collection, model training, evaluation, and online serving.
    In the Xiaomi Home app, a privacy agreement pop-up is shown when the smart home command understanding feature is first launched. The functionality is enabled only after explicit user consent, and users can opt out at any time via the settings page.
    When we synthesize training and evaluation datasets, all identifier fields (e.g., user IDs, device IDs) are strictly anonymized. Potentially privacy-related text (e.g., device aliases) is de-identified via normalization and replacement with abstract placeholders prior to training and evaluation.
    Finally, we enforce data minimization and governance by limiting data usage to what is necessary for model development and quality assessment, and restricting access to authorized personnel under standard security controls.

\section{Limitation}
    A primary limitation of the current MiCU system is its reliance on single-turn command understanding, without incorporating long-term user interaction history. Smart home environments are inherently personal; however, our model treats each request as an independent event. This lack of historical context prevents the system from learning individual user habits or "personality," such as a user’s specific preference for "cozy" lighting at different times of the day. Consequently, the model may struggle to resolve extreme ambiguities that require personalized knowledge, which we identify as a key direction for future research.

\end{document}